\documentclass[journal]{IEEEtran}
\usepackage{amsmath,amssymb}
\usepackage{graphicx}
\usepackage{booktabs}
\usepackage{multirow}
\usepackage{algorithm}
\usepackage{algpseudocode}
\usepackage{url}
\usepackage[hidelinks]{hyperref}

\DeclareMathOperator*{\E}{\mathbb{E}}

\begin{document}

\title{RBMatch: Dual-Level Class Rebalancing for\\
Semi-Supervised Building Footprint Extraction}

\author{Akil~Ahmad~Taki\textsuperscript{1,2}, Shaikh~Anowarul~Fattah\textsuperscript{1}~\IEEEmembership{~Member,~IEEE}%
\thanks{\textsuperscript{1} Department of EEE, Bangladesh University of Engineering and Technology(BUET),\textsuperscript{2}Department of EEE, University of Asia Pacific 
(e-mail: 0424062543@eee.buet.ac.bd, akil@uap-bd.edu, fattah@eee.buet.ac.bd} }


\maketitle

\begin{abstract}
Accurate building footprint extraction from high-resolution remote sensing imagery underpins urban planning, disaster response, and environmental monitoring. This remains a challenge, as conventional supervised learning involves extensive pixel-wise annotation. On the other hand, Semi-supervised learning (SSL) reduces this burden by exploiting unlabeled data through self-training. However, remote sensing images are inherently dominated by background pixels, and this class imbalance biases the unsupervised gradient toward the majority class during self-training. We found that fixing this imbalance at a single stage falls short. If only the pseudo-labels are balanced, the background bias still reappears during the unsupervised loss calculation, an issue we term the \emph{imbalance leak.} To address this issue, we propose RBMatch, a dual-level class-rebalancing framework that intervenes at both the pseudo-label and loss levels of self-training. RBMatch integrates a standard supervised pathway with a robust self-training module that incorporates three components: adaptive class-specific thresholding (ACT) for balanced pseudo-label selection, confidence-aware class-balanced reweighting (CACBR) of the unsupervised loss, and a distribution alignment layer (DAL) that matches predicted class distributions on unlabeled data to the labeled-data prior. Experiments on the WHU, INRIA, and Massachusetts building datasets under labeled ratios of 1\%--10\% show that RBMatch attains the highest building IoU and F1 in every setting evaluated. Gains are largest on the most imbalanced benchmark, Massachusetts, where RBMatch improves IoU by $1.37$ over the strongest baseline at $r\!=\!0.01$ and it is the only method to exceed the supervised baseline in all twelve dataset--ratio settings.
\end{abstract}


\section{Introduction}
\IEEEPARstart{B}{uilding} footprint extraction (pixel-level segmentation of buildings from optical imagery) supports urban-growth monitoring, damage assessment, and population estimation. Transformers like SegFormer~\cite{segformer} set strong benchmarks but require expensive dense pixel-wise annotations. Semi-supervised learning (SSL) addresses this bottleneck by pairing a small labeled set with a large unlabeled corpus. The dominant paradigm is weak-to-strong self-training: pseudo-labels from weakly augmented unlabeled images supervise predictions on strongly augmented counterparts~\cite{fixmatch, unimatch}. Self-training pipelines divide into consistency-based~\cite{cct,cps} and pseudo-label-based methods~\cite{fixmatch}, with the latter dominating recent state-of-the-art.

This paradigm exposes a structural difficulty on building data: \emph{severe class imbalance}. Building pixels occupy at most 10--25\% of image area, and as little as 3--5\% in suburban scenes. This imbalance creates a positive feedback loop: initial background bias yields skewed pseudo-labels, which reinforces the bias during training---a form of confirmation bias~\cite{arazo}.

Recent work corrects imbalance at distinct pipeline stages. At the logit level, DBMatch~\cite{dbmatch} subtracts a global bias estimated from class-conditional logit means. At the threshold level, FreeMatch~\cite{freematch} and AdaptMatch~\cite{adaptmatch} replace FixMatch's fixed $\tau=0.95$ with dynamic thresholds that admit more minority pseudo-labels. Class imbalance is also treated at the loss level via focal loss~\cite{focal}, effective-number balancing~\cite{cb}, margin-aware losses~\cite{ldam,logitadj}, distribution alignment~\cite{remixmatch}, and labeled-data-anchored resampling (DARS~\cite{dars}). In remote-sensing SSL, Zhang \emph{et al.}~\cite{zhangjoint} rebalance which samples enter consistency regularization at the input-batch level.

These methods improve on FixMatch but share a structural flaw: \emph{relying on a single corrective intervention.} After any threshold correction, the selected set $S$ remains dominated by background due to spatial sparsity (background pixels vastly outnumber building pixels) and confidence asymmetry (homogeneous background accumulates high-confidence pixels quickly). Because the unsupervised loss aggregates per-pixel terms uniformly, the gradient stays background-dominated regardless of how $S$ is constructed. Threshold rebalancing fixes \emph{which} pseudo-labels are used; it does not fix \emph{how much} each contributes. We term this residual skew the \emph{imbalance leak}.

To address these challenges, the main contributions are summarized as follows:
\begin{itemize}
    \item We empirically characterize the imbalance leak, showing that residual selection skew drives majority-class gradient dominance.
    \item We propose RBMatch, the first remote-sensing SSL framework to coordinate rebalancing at \emph{both} the pseudo-label level (via adaptive thresholding and distribution alignment) and loss level (via confidence-aware reweighting).
    \item We demonstrate that RBMatch achieves state-of-the-art building IoU/F1 across three benchmarks at diverse labeled ratios.
    \item We provide mechanistic evidence via gradient-weight logging that our loss-level intervention mitigates residual pseudo-label imbalance.
\end{itemize}

\begin{figure*}[t]
\centering
\includegraphics[width=0.98\textwidth]{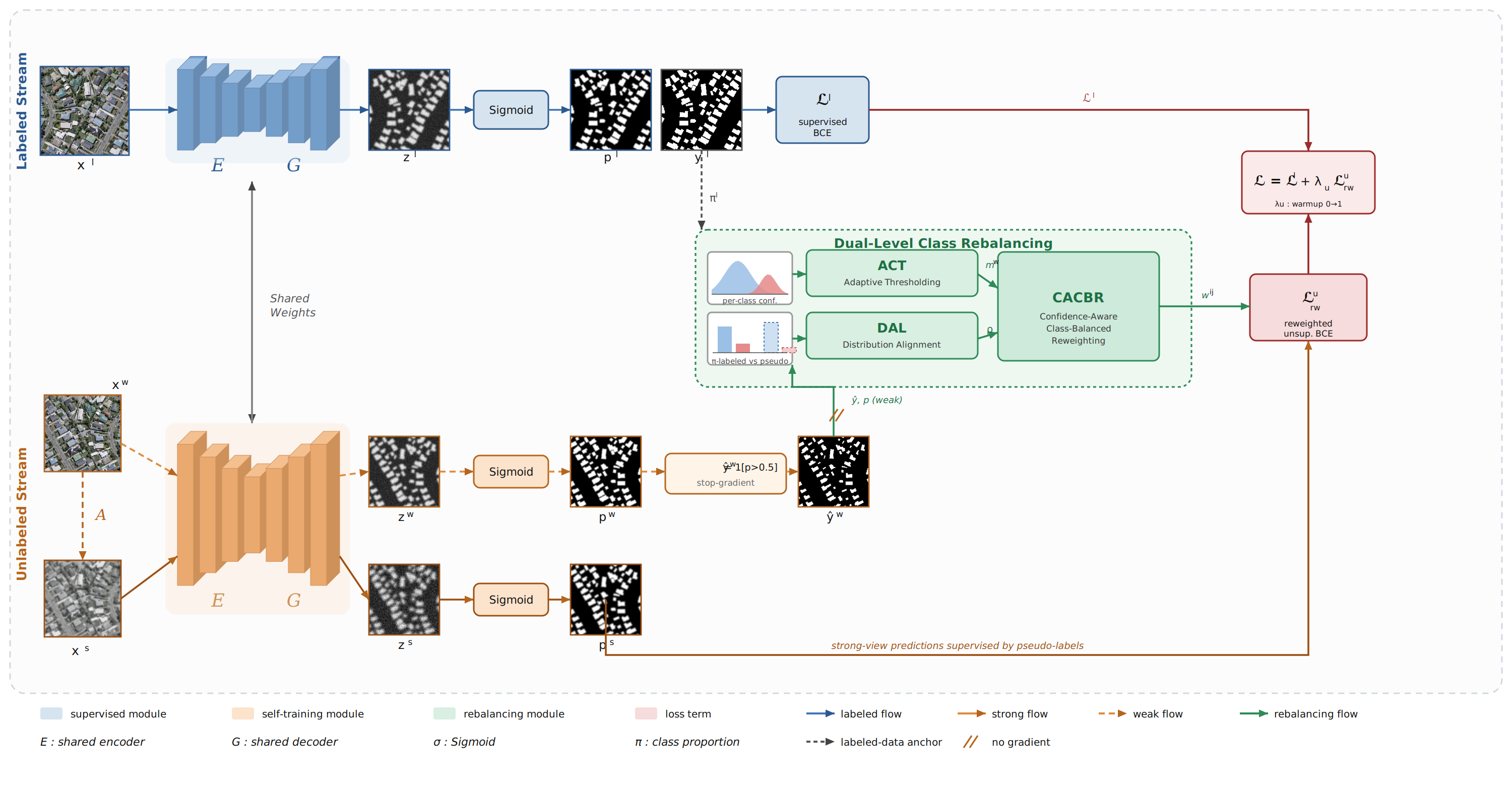}
\caption{Proposed RBMatch architecture. A shared SegFormer-B2 backbone processes the labeled stream and the weak/strong unlabeled views. Weak-view pseudo-labels (gradient stopped) drive ACT (adaptive thresholding), DAL (distribution alignment, anchored by $\pi^l$), and CACBR (confidence-aware reweighting), which produce per-pixel weights $w_{ij}$ for the reweighted unsupervised loss on the strong view.}
\label{fig:workflow}
\vspace{-1em}
\end{figure*}

\section{Preliminaries}
Let $\mathcal{D}_l=\{(x_n^l,y_n^l)\}_{n=1}^{N_l}$ be the labeled set, with $x_n^l\!\in\!\mathbb{R}^{H\times W\times3}$ and pixel-wise binary labels $y_n^l\!\in\!\{0,1\}^{H\times W}$ ($0$\,=\,background, $1$\,=\,building), and $\mathcal{D}_u=\{x_m^u\}_{m=1}^{N_u}$ the unlabeled set with $N_u\!\gg\!N_l$. The labeled ratio is $r=N_l/(N_l+N_u)$; we report $r\in\{0.01,0.02,0.05,0.10\}$. Following~\cite{dbmatch}, all methods use SegFormer-B2 as a shared encoder--decoder backbone producing a logit map $z=G(E(x))\in\mathbb{R}^{H\times W}$ and probability map $p_{ij}=\sigma(z_{ij})$. For each unlabeled image, we form a weakly augmented view $x^w$ (flip, scale in $[0.5,2.0]$, crop) and a strongly augmented view $x^s=\mathcal{A}_s(x^w)$ (color jitter, grayscale, blur)~\cite{unimatch,dbmatch}.

\section{The RBMatch Framework}
Imbalance in SSL building extraction is a multi-stage phenomenon. Closing the leak requires acting where it is created (selection) and where it operates (the gradient). Three hypotheses frame our modules: 

\begin{itemize}

\item \textbf{H1 (Multi-stage imbalance):} Effective correction requires intervention at both pseudo-label selection (\emph{which} pixels participate) and loss computation (\emph{how much} each contributes). \item \textbf{H2 (Confidence-aware reweighting):} Loss-level rebalancing must be modulated by per-pixel confidence to amplify high-quality minority pseudo-labels rather than indiscriminately boosting all minority pixels. 
\item\textbf{H3 (Labeled-data anchoring):} Reweighting intensity must be anchored to the labeled-data class distribution to prevent self-reinforcing drift.
\end{itemize}

RBMatch builds on the FixMatch template (Fig.~\ref{fig:workflow}). The labeled stream produces the supervised loss; the unlabeled image is augmented weakly to $x^w$ (whose pseudo-labels $\hat y^w=\mathbf{1}[p^w>0.5]$ are formed with stopped gradients) and strongly to $x^s$. Three modules operate on the weak-view pseudo-labels: ACT produces a class-balanced selection mask, DAL computes a drift coefficient anchored to the labeled-data proportions, and CACBR converts these into per-pixel weights $w_{ij}$ for the reweighted unsupervised loss.

\emph{Supervised branch.} For a labeled pair, the supervised loss is the pixel-averaged BCE:
\begin{equation}
\mathcal{L}_l=\frac{1}{HW}\sum_{i,j}\!\big[-y_{ij}^l\log p_{ij}^l-(1-y_{ij}^l)\log(1-p_{ij}^l)\big].
\label{eq:sup}
\end{equation}
This branch yields the frozen reference proportions $\pi_c^l=\big(\sum_{n}\sum_{i,j}\mathbf{1}[y_{n,ij}^l=c]\big)/(N_l HW)$, $c\in\{0,1\}$, computed once over $\mathcal{D}_l$; DAL uses $\pi^l$ to anchor rebalancing.

\subsection{Adaptive Class-Specific Thresholding (ACT)}
FixMatch's fixed $0.95$ threshold starves minority pseudo-labels early in training. Following FreeMatch~\cite{freematch}, ACT tracks the model's per-class confidence on the unlabeled stream by EMA,
\begin{equation}
\hat\tau_c\leftarrow\alpha_\tau\hat\tau_c+(1-\alpha_\tau)\,\E_{(i,j)\in B_c}\!\big[p_{ij}^w\big],
\label{eq:act}
\end{equation}
with $B_c=\{(i,j):\hat y_{ij}^w=c\}$ and $\alpha_\tau=0.999$, and forms the class-conditional selection mask:
\begin{equation}
m_{ij}^w=\mathbf{1}\!\big[\,p_{ij}^w>\hat\tau_1\ \lor\ p_{ij}^w<1-\hat\tau_0\,\big].
\label{eq:mask}
\end{equation}
When the model is uncertain about buildings, \eqref{eq:act} drives $\hat\tau_1$ down, letting more building pseudo-labels through. The property $\hat\tau_1<\hat\tau_0$ emerges structurally from EMA tracking.

\subsection{Confidence-Aware Class-Balanced Reweighting (CACBR)}
ACT fixes \emph{which} pseudo-labels enter the loss, but the surviving set remains background-dominated. CACBR replaces uniform aggregation with a per-pixel weight multiplying a class-balance factor, a confidence factor, and the ACT mask: $w_{ij}=w_{\mathrm{class}}(\hat y_{ij}^w)\cdot w_{\mathrm{conf}}(p_{ij}^w)\cdot m_{ij}^w$. The class-balance factor uses the effective-number formulation~\cite{cb},
\begin{equation}
w_{\mathrm{class}}(c)=\frac{1-\beta}{1-\beta^{\,n_c}},
\label{eq:cb}
\end{equation}
with $n_c$ the EMA-tracked count of selected class-$c$ pixels and $\beta=0.9999$, normalized so $\sum_c\pi_c^l w_{\mathrm{class}}(c)=1$. The confidence factor modulates by per-pixel confidence: $w_{\mathrm{conf}}=(p_{ij}^w)^{\gamma}$ if $\hat y_{ij}^w=1$ and $(1-p_{ij}^w)^{\gamma}$ if $\hat y_{ij}^w=0$, with $\gamma=1$. The multiplicative form ensures minority pixels require high confidence to receive large weights. The reweighted unsupervised loss aggregates by weight:
\begin{equation}
\mathcal{L}_{\mathrm{rw}}^{u}=\frac{\sum_{i,j}w_{ij}\,\mathrm{BCE}(p_{ij}^s,\hat y_{ij}^w)}{\sum_{i,j}w_{ij}+\epsilon}, \quad \epsilon=10^{-8}.
\label{eq:lrw}
\end{equation}

\subsection{Distribution Alignment Layer (DAL)}
If early pseudo-labels are background-skewed, CACBR amplifies the minority class based on a corrupted estimate. DAL guards against this loop by anchoring to the trustworthy labeled distribution $\pi^l$ (H3). Let $\hat\pi_c^u$ be the EMA-tracked pseudo-label proportion ($\alpha_\pi=0.999$). The drift coefficient $\rho$ scales the minority weight asymmetrically, leaving the background untouched:

\begin{equation}
\rho=\min\!\Big(\frac{\pi_1^l/\hat\pi_1^u}{\pi_0^l/\hat\pi_0^u}, \rho_{\max}\Big), \quad \tilde w_{\mathrm{class}}(1)=\rho\,w_{\mathrm{class}}(1),
\label{eq:rho}
\end{equation}
and $\tilde w_{\mathrm{class}}(0)=w_{\mathrm{class}}(0)$. DAL is strictly a minority-class booster; the cap $\rho_{\max}=3.0$ prevents transient EMA instability from triggering excessive amplification. The final weight is $w_{ij}=\tilde w_{\mathrm{class}}(\hat y_{ij}^w)\,w_{\mathrm{conf}}(p_{ij}^w)\,m_{ij}^w$. 

\emph{Overall objective.} The total loss combines both terms:
\begin{equation}
\mathcal{L}=\mathcal{L}_l+\lambda_u\,\mathcal{L}_{\mathrm{rw}}^{u},
\label{eq:total}
\end{equation}
with $\lambda_u$ warmed linearly from $0$ to $1$ over $500$ iterations. Algorithm~\ref{alg} summarizes one iteration. In contrast to DBMatch (global scalar logit shift), FreeMatch (selection-only), and DARS (resampling), RBMatch operates at a per-pixel granularity across multiple training stages.

\begin{algorithm}[h]
\caption{RBMatch training loop (one iteration).}
\label{alg}
\textbf{Input:} Labeled batch $\{(x^l,y^l)\}$, unlabeled batch $\{x^u\}$, model $E\!\circ\!G$, EMA state $(\hat\tau_0,\hat\tau_1,n_0,n_1,\hat\pi_0^u,\hat\pi_1^u)$, frozen $\pi^l$, hyperparameters.\\
\textbf{Output:} Updated parameters and EMA state.
\begin{algorithmic}[1]
\State Forward pass: compute $p^l,p^w,p^s$ through shared backbone
\State Supervised loss $\mathcal{L}_l$ via \eqref{eq:sup}; Pseudo-labels $\hat y_{ij}^w\!\leftarrow\!\mathbf{1}[p_{ij}^w>0.5]$
\State ACT: update $\hat\tau_c$ via \eqref{eq:act}; mask $m^w$ via \eqref{eq:mask}
\State DAL: update $\hat\pi_c^u$; drift $\rho$ via \eqref{eq:rho}
\State CACBR: update $n_c, w_{\mathrm{class}}$; DAL scale \eqref{eq:rho}; weights $w_{ij}$
\State Reweighted loss $\mathcal{L}_{\mathrm{rw}}^{u}$ via \eqref{eq:lrw}; Backprop $\mathcal{L}\!\leftarrow\!\mathcal{L}_l+\lambda_u\mathcal{L}_{\mathrm{rw}}^{u}$
\end{algorithmic}
\end{algorithm}

\section{Experiments}
\subsection{Setup}
We evaluate on the WHU Building Dataset~\cite{whu} (official 5\,733/1\,228/1\,228 splits), the INRIA Vienna/Tyrol/Kitsap subsets~\cite{inria}, and the Massachusetts Boston subset~\cite{mass}, cropped to $512\times512$~\cite{dbmatch}. The WHU labeled distribution is $\pi_0^l\!=\!0.841,\pi_1^l\!=\!0.159$ at $r\!=\!0.01$. Baselines include Only-Sup, FixMatch~\cite{fixmatch}, DBMatch~\cite{dbmatch}, FreeMatch~\cite{freematch}, AdaptMatch~\cite{adaptmatch}, and DARS~\cite{dars}. We report Recall, Precision, IoU, and F1 for the building class. Methods utilize SegFormer-B2 with identical augmentations, AdamW (lr $2.5\times10^{-4}$, poly decay $0.9$), batch sizes 4/4, and $500$-iteration warmup. RBMatch uses $\alpha_\tau\!=\!\alpha_\pi\!=\!0.999$, $\beta\!=\!0.9999$, $\gamma\!=\!1$, $\rho_{\max}\!=\!3.0$, and $\lambda_u\!=\!1$. Experiments run on an RTX-3090 GPU; our modules add under 5\% forward-pass time. 

\subsection{Diagnostic: demonstrating the imbalance leak}
\begin{figure}[h]
\centering
\includegraphics[width=\linewidth]{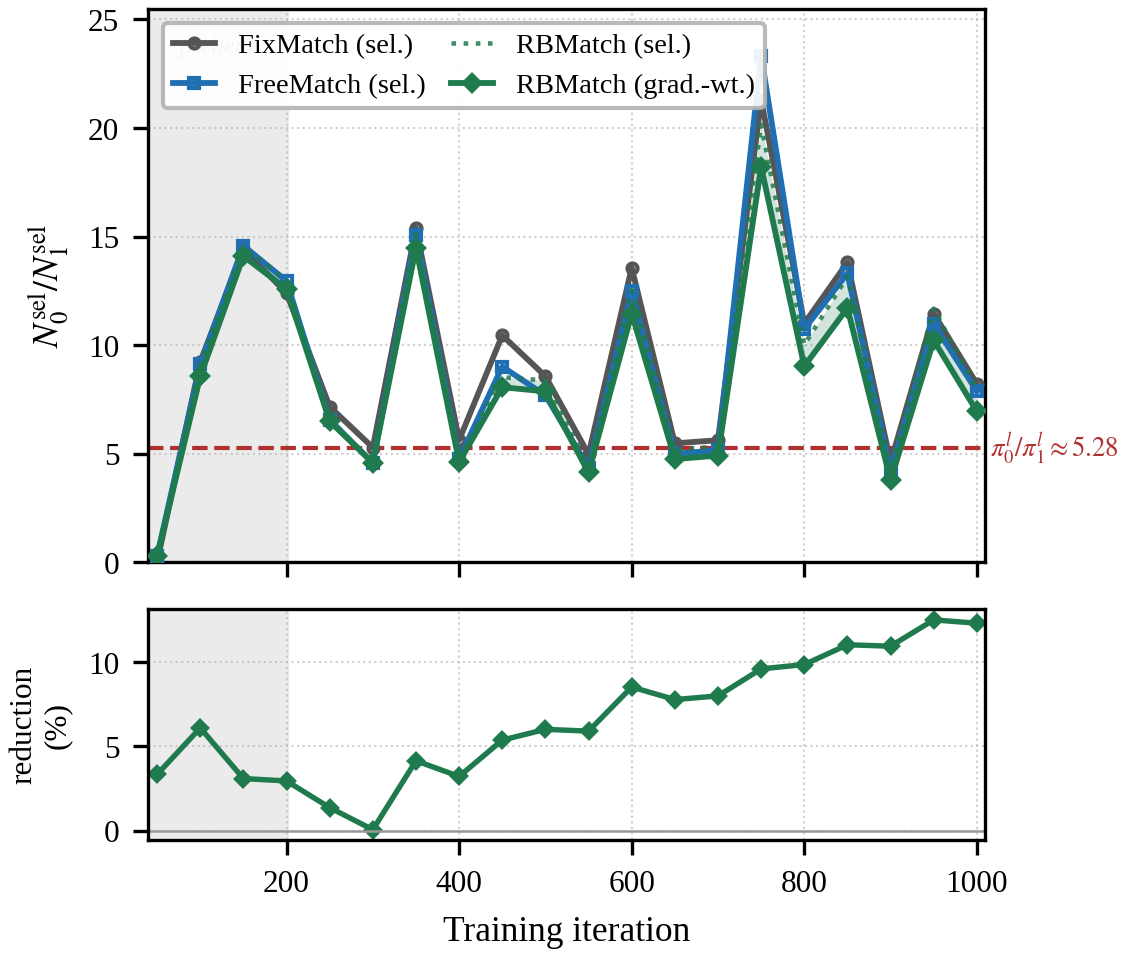}
\caption{Empirical demonstration of the imbalance leak on WHU ($r\!=\!0.01$). RBMatch's gradient-weighted effective ratio corrects FreeMatch/FixMatch bias.}
\label{fig:leak}
\vspace{-1em}
\end{figure}

We log the selection ratio $N_0^{\mathrm{sel}}/N_1^{\mathrm{sel}}$ and gradient-weighted effective ratio $\big(\sum_{\hat y^w=0}w_{ij}\big)/\big(\sum_{\hat y^w=1}w_{ij}\big)$ every $50$ iterations. On WHU ($r\!=\!0.01$), FixMatch oscillates between $\sim$5--28, spiking on sparse batches. DBMatch performs similarly because its global logit shift does not alter pixel selection. FreeMatch lowers the ratio to $\sim$8--15, confirming selection rebalancing leaves a residual leak. RBMatch's gradient-weighted ratio remains strictly below the raw selection ratio across all iterations (Fig.~\ref{fig:leak}). By iteration $750$, the raw ratio is $9.69$ versus a gradient-weighted $\approx7$. This expanding gap supports H1, driving the effective gradient closer to the true distribution.

\begin{figure*}[t]
\centering
\includegraphics[width=\textwidth]{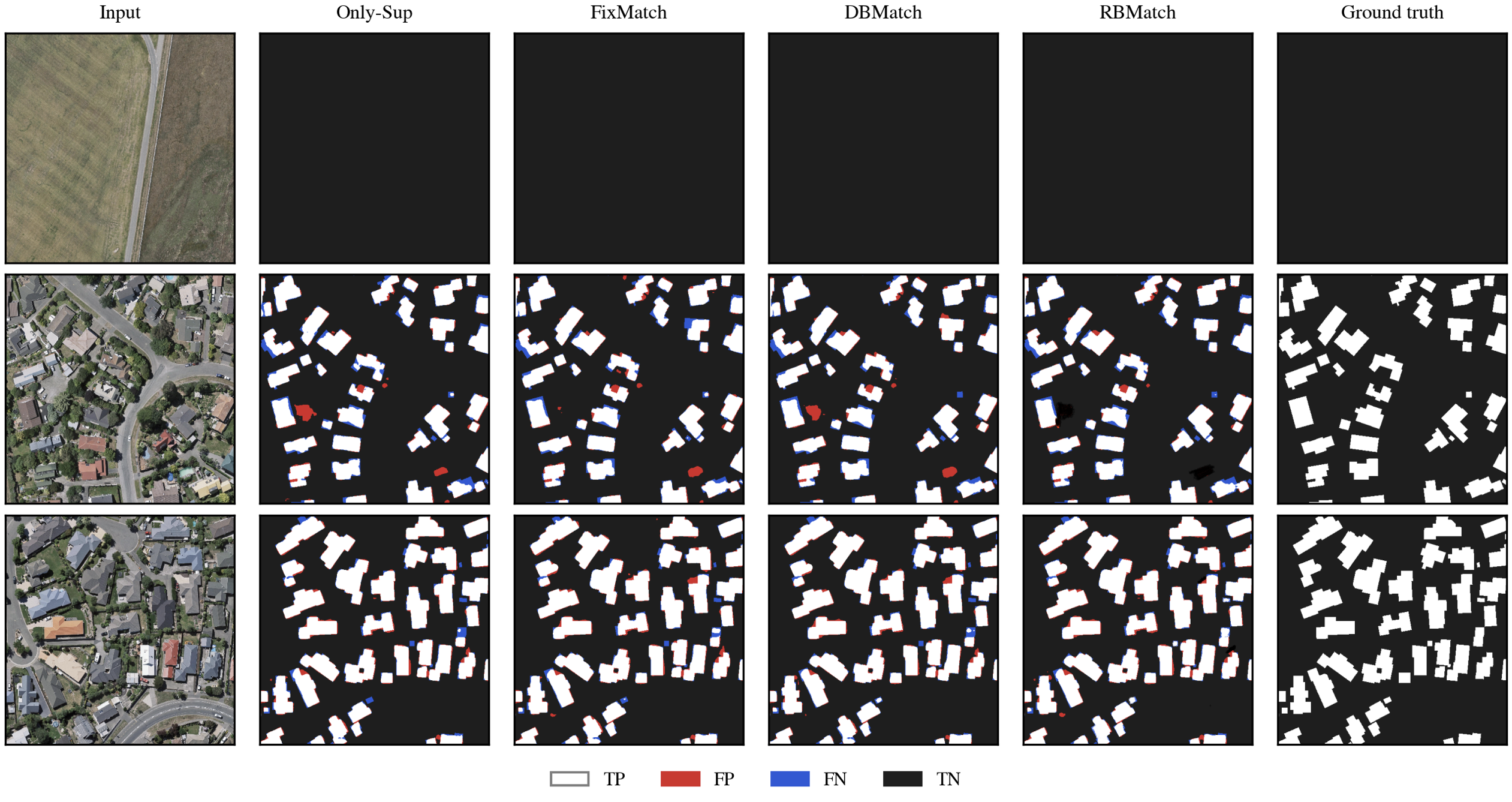}
\caption{Qualitative comparison on WHU ($r\!=\!0.01$). White: TP, red: FP, blue: FN, black: TN. RBMatch recovers complete interiors without inflating false positives.}
\label{fig:qual}
\vspace{-1em}
\end{figure*}

\subsection{Ablation}
Six configurations evaluated on WHU at $r\!=\!0.05$
(Table~\ref{tab:abl}) isolate the contribution of each module.
Naive self-training (b) underperforms the supervised baseline (a)
by $1.23$ IoU, consistent with confirmation bias: unchecked
background-dominated pseudo-labels reinforce the majority-class
prior. ACT alone (c) and CACBR alone (d) each recover roughly
$60\%$ of this deficit, but neither restores the baseline. Their
near-identical results ($83.89$ vs.\ $83.87$ IoU) indicate that
the two interventions act on largely overlapping sources of skew
when applied in isolation. Combining them without DAL (e) yields
$84.33$ IoU and still falls short of Only-Sup, because the
reweighting remains tied to uncalibrated model predictions rather
than to a trustworthy reference distribution. The full framework
(f) adds DAL to anchor the weights to $\pi^l$, reaching $85.19$
IoU and establishing the only configuration that clears the
supervised baseline---validating H3. The gain is driven by
recall, which improves from $88.55$ (b) to $90.21$ while precision
is preserved ($93.14$ to $93.88$), confirming that dual-level
rebalancing recovers minority-class pixels without inflating false
positives.

\begin{table}[h]
\centering
\caption{Ablation on WHU at $r\!=\!0.05$. Best in \textbf{bold}.}
\label{tab:abl}
\footnotesize
\setlength{\tabcolsep}{4pt}
\begin{tabular}{@{}llcccc@{}}
\toprule
 & Configuration & Recall & Precision & IoU & F1 \\
\midrule
(a) & Only-Sup            & 89.68 & 93.43 & 84.36 & 91.52 \\
(b) & FixMatch (SUP+ST)   & 88.55 & 93.14 & 83.13 & 90.79 \\
(c) & \quad + ACT only    & 88.60 & 94.04 & 83.89 & 91.24 \\
(d) & \quad + CACBR only  & 88.58 & 94.04 & 83.87 & 91.23 \\
(e) & ACT+CACBR (no DAL)  & 89.31 & 93.80 & 84.33 & 91.50 \\
(f) & \textbf{RBMatch (full)} & \textbf{90.21} & \textbf{93.88} & \textbf{85.19} & \textbf{92.00} \\
\bottomrule
\end{tabular}
\end{table}

\subsection{Comparison with state of the art}
Table~\ref{tab:benchmarks} presents quantitative results. RBMatch secures the highest IoU and F1 score at every labeled ratio across all datasets. On WHU, its performance edge over DBMatch expands alongside the label budget ($+0.04$ at $r\!=\!0.01$ to $+1.97$ at $r\!=\!0.10$). On INRIA, RBMatch dominates every column while baselines alternate second-best status. The most pronounced margins occur on Massachusetts—the most imbalanced benchmark—yielding a $+1.37$ IoU improvement over FreeMatch at $r\!=\!0.01$. Crucially, across all twelve dataset--ratio settings RBMatch is the
only method that never falls below the supervised baseline, while every competing SSL method regresses in at least two settings.

\begin{table}[h]
\centering
\caption{IoU/F1 (\%) across datasets. Best in \textbf{bold}; second-best \underline{underlined}.}
\label{tab:benchmarks}
\footnotesize
\setlength{\tabcolsep}{3pt}
\begin{tabular}{@{}lcccc@{}}
\toprule
Method & $r$=0.01 & $r$=0.02 & $r$=0.05 & $r$=0.10 \\
\midrule
\multicolumn{5}{c}{\textbf{WHU Building Dataset}} \\
\midrule
Only-Sup   & 83.27/90.87 & 83.55/91.04 & \underline{84.36}/\underline{91.52} & \underline{85.38}/\underline{92.11} \\
FixMatch   & 82.42/90.36 & 82.89/90.64 & 83.13/90.79 & 84.74/91.74 \\
FreeMatch  & 81.79/89.98 & 82.05/90.14 & 82.80/90.59 & 83.94/91.27 \\
AdaptMatch & 82.62/90.48 & 83.21/90.84 & 83.31/90.90 & 84.88/91.82 \\
DARS       & 82.55/90.44 & 83.41/90.95 & 83.16/90.81 & 84.43/91.56 \\
DBMatch    & \underline{83.48}/\underline{91.00} & \underline{84.01}/\underline{91.31} & 83.90/91.25 & 85.23/92.03 \\
\textbf{RBMatch} & \textbf{83.52}/\textbf{91.02} & \textbf{84.30}/\textbf{91.48} & \textbf{85.19}/\textbf{92.00} & \textbf{87.20}/\textbf{93.16} \\
\midrule
\multicolumn{5}{c}{\textbf{INRIA Aerial Image Labeling}} \\
\midrule
Only-Sup   & 68.76/81.49 & \underline{71.41}/\underline{83.32} & 70.94/83.00 & 71.32/83.26 \\
FreeMatch  & 69.47/81.99 & 70.96/83.01 & \underline{71.83}/\underline{83.61} & 71.17/83.16 \\
AdaptMatch & 69.80/82.21 & 71.08/83.09 & 71.50/83.38 & \underline{72.45}/\underline{84.02} \\
DBMatch    & \underline{69.94}/\underline{82.31} & 68.17/81.07 & 70.39/82.62 & 71.55/83.42 \\
\textbf{RBMatch} & \textbf{70.96}/\textbf{83.01} & \textbf{72.32}/\textbf{83.94} & \textbf{72.39}/\textbf{83.98} & \textbf{72.88}/\textbf{84.31} \\
\midrule
\multicolumn{5}{c}{\textbf{Massachusetts Boston Subset}} \\
\midrule
Only-Sup   & 53.66/69.84 & 56.91/72.54 & 62.07/76.60 & 63.06/77.35 \\
FreeMatch  & \underline{57.29}/\underline{72.85} & 58.23/73.60 & 62.59/76.99 & 62.97/77.28 \\
AdaptMatch & 52.84/69.14 & 57.63/73.12 & \underline{62.95}/\underline{77.26} & 63.31/77.53 \\
DBMatch    & 54.10/70.21 & \underline{58.61}/\underline{73.90} & 62.62/77.01 & \underline{63.46}/\underline{77.65} \\
\textbf{RBMatch} & \textbf{58.66}/\textbf{73.94} & \textbf{60.68}/\textbf{75.53} & \textbf{63.95}/\textbf{78.01} & \textbf{64.06}/\textbf{78.09} \\
\bottomrule
\end{tabular}
\end{table}

\subsection{Generality across backbones}
To verify that RBMatch is not tied to transformer-specific
behaviour, we evaluate it across UNet~\cite{unet}, UNet++, and
DeepLabV3 on WHU at $r\!=\!0.05$ (Table~\ref{tab:backbone}).
RBMatch attains the best IoU and F1 on every backbone. Its margin
over DBMatch is stable across architecture families
($+1.81$, $+1.54$, $+1.69$, and $+1.29$ IoU on UNet, UNet++,
DeepLabV3, and SegFormer-B2, respectively), indicating that
dual-level rebalancing acts on the pseudo-label distribution
rather than on any property of a particular encoder.

The comparison against the supervised baseline is more
discriminative. DBMatch improves over Only-Sup on the three CNNs
($+2.22$ to $+0.33$ IoU) but falls below it on SegFormer-B2
($83.90$ vs.\ $84.36$), whereas RBMatch clears the baseline on all
four backbones ($+4.03$, $+4.00$, $+2.02$, and $+0.83$ IoU).
Correcting the imbalance leak at the loss level therefore
determines whether self-training helps or hurts, which supports
H1. 

\begin{table}[h]
\centering
\caption{Generality across backbones on WHU at $r\!=\!0.05$. Best in bold.}
\label{tab:backbone}
\footnotesize
\setlength{\tabcolsep}{3pt}
\begin{tabular}{@{}lcccc@{}}
\toprule
Method & UNet & UNet++ & DeepLabV3 & SegFormer-B2 \\
\midrule
Only-Sup & 73.16/84.50 & 75.37/85.95 & 78.18/87.75 & 84.36/91.52 \\
DBMatch  & 75.38/85.96 & 77.83/87.53 & 78.51/87.96 & 83.90/91.25 \\
\textbf{RBMatch} & \textbf{77.19/87.13} & \textbf{79.37/88.50} & \textbf{80.20/89.01} & \textbf{85.19/92.00} \\
\bottomrule
\end{tabular}
\end{table}

\subsection{Qualitative analysis, discussion, and limitations}
\label{sec:disc}

Fig.~\ref{fig:qual} illustrates model outputs. RBMatch resolves false-negative errors inside building structures, leading to solid interior segments without introducing false-positive artifacts, matching expectations for joint threshold and loss adjustments. DBMatch edits show uniform changes along edges. The framework yields optimal increments under skewed profiles and severe data limitations, satisfying H1. Limitations: Expansion to multi-class scenarios requires multi-dimensional alignment; label noise in the reference ground-truth $\pi^l$ propagates into DAL anchors.

\section{Conclusion}
We introduced RBMatch, a dual-level class-rebalancing framework for semi-supervised building footprint extraction. RBMatch closes the imbalance leak with three coordinated modules (ACT, CACBR, DAL) acting at the threshold and loss levels. It attains state-of-the-art building IoU/F1 among evaluated SSL methods on WHU, INRIA, and Massachusetts, with the largest margins on the most imbalanced benchmarks. Future work includes multi-class extensions and SAM-based~\cite{sam} boundary refinement.

\end{document}